\documentclass[letterpaper, 10 pt, journal, twoside]{IEEEtran}
\ifCLASSINFOpdf
\else
\fi
\usepackage{graphics}
\usepackage{epsfig}
\usepackage{amssymb}
\usepackage[T1]{fontenc}
\usepackage{xcolor}
\usepackage{amsmath}
\usepackage{booktabs}
\usepackage{hyperref}
\usepackage{amsfonts}
\usepackage{marvosym}
\begin{document}
%
\title{Achieving Stable High-Speed Locomotion for Humanoid Robots with Deep Reinforcement Learning}
%
%
%

\author{Xinming Zhang\textsuperscript{1, 2\dag}, Xianghui Wang\textsuperscript{2\dag}, Lerong Zhang\textsuperscript{2}, Guodong Guo\textsuperscript{1, 2}, Xiaoyu Shen\textsuperscript{2} and Wei Zhang\textsuperscript{2}%


\thanks{This work has been submitted to the IEEE for possible publication. Copyright may be transferred without notice, after which this version may no longer be accessible.}

\thanks{This work is supported by 2035 Key Research and Development Program of Ningbo City under Grant No.2024Z127. \textit{(Corresponding author: Wei Zhang.)}}

\thanks{$^{1}$Xinming Zhang and Guodong Guo are with the School of Computer Science and Technology, University of Science and Technology of China, Hefei, China
        {\tt\footnotesize (e-mail: xm\_zhang@mail.ustc.edu.cn; gdguo@eitech.edu.cn).}}%
\thanks{$^{2}$Xinming Zhang, Xianghui Wang, Lerong Zhang, Guodong Guo, Xiaoyu Shen and Wei Zhang are with the Ningbo Institute of Digital Twin, Eastern Institute of Technology, Ningbo, China
        {\tt\footnotesize (e-mail: xm\_zhang@mail.ustc.edu.cn; xhwang@eitech.edu.cn; zhanglr7@outlook.com; gdguo@eitech.edu.cn; xyshen@eitech.edu.cn; zhw@eitech.edu.cn).}}%
\thanks{$^{\dag}$Xinming Zhang and Xianghui Wang contributed equally to this work as co-first authors.}%
}
\maketitle

\begin{abstract}
Humanoid robots offer significant versatility for performing a wide range of tasks, yet their basic ability to walk and run, especially at high velocities, remains a challenge. This letter presents a novel method that combines deep reinforcement learning with kinodynamic priors to achieve stable locomotion control (KSLC). KSLC promotes coordinated arm movements to counteract destabilizing forces, enhancing overall stability. Compared to the baseline method, KSLC provides more accurate tracking of commanded velocities and better generalization in velocity control. In simulation tests, the KSLC-enabled humanoid robot successfully tracked a target velocity of 3.5 m/s with reduced fluctuations. Sim-to-sim validation in a high-fidelity environment further confirmed its robust performance, highlighting its potential for real-world applications.
\end{abstract}

\begin{IEEEkeywords}
Humanoid and Bipedal Locomotion, Humanoid Robot Systems, Machine Learning for Robot Control
\end{IEEEkeywords}

%
\IEEEpeerreviewmaketitle

\section{Introduction}
%
%
%
%

\IEEEPARstart{I}{n} human-centered environments, the human-like skeletal structure of humanoid robots allows for better maneuverability in complex terrains and human-designed buildings. This enables them to perform a wider range of tasks compared to wheeled robots by utilizing existing human infrastructure. The locomotion capacity of humanoid robots is crucial for performing multiple tasks in the real world. However, applying classical control methods to humanoid robot locomotion tasks requires considerable manual effort to handle specific situations. Deep Reinforcement Learning (DRL)-based methods, on the other hand, derive optimal control strategies through automatic iterative trial-and-error in simulated environments \cite{kang2023rl+}, resulting in greater robustness and less manual design effort. Notably, parallel training for DRL can significantly reduce the required training time, while reducing hardware costs and improving computational efficiency. Consequently, DRL has emerged as a popular methodology in the field of robotics control \cite{rudin2022learning, hoeller2024anymal, margolis2024rapid}.

Despite advances in humanoid robotics, controlling locomotion remains a substantial challenge due to the complexity of their larger action space, which often leads to instability \cite{cheng2024expressive}. While DRL-based methods have shown remarkable progress, their success has primarily been limited to low-velocity locomotion (approximately 1.0 m/s) \cite{radosavovic2024real, gu2024humanoid}, and typically with fewer actuated joints. When upper-body joints are fully engaged, the robot's movements can become unpredictable, resulting in unnatural and erratic twisting behavior \cite{sferrazza2024humanoidbench}. \textbf{In this letter, we challenge the conventional approach of disabling the arm joints during locomotion training \cite{gu2024humanoid, sferrazza2024humanoidbench}. Instead, we demonstrate that leveraging the arm joints can substantially improve the effectiveness of training for both walking and running.}

At high speeds, humanoid robots frequently fall due to imbalanced angular momentum, primarily caused by the swinging leg in the global yaw direction \cite{otani2018upper}. The primary goal of this work is to achieve precise and stable locomotion control, preventing the development of a wobbly gait or falls.

To achieve this, we develop a novel method called Kinodynamic-regularized Stable Locomotion Control (KSLC), a DRL-based approach for stabilizing humanoid robots. The KSLC integrates DRL with prior kinodynamic knowledge to discover the optimal control policy. Specifically, we design a reward function that minimizes angular momentum \cite{raibert1986legged}, encouraging the robot to move forward using its legs while generating corresponding arm movements and elbow flexion to help stabilize its body in the global yaw direction.

Additionally, we observed suboptimal velocity-tracking performance in the humanoid robot across a wide range of commanded velocities when fixed rewards were used. This limitation arises because fixed rewards tend to emphasize a single motion pattern. To improve the robot's velocity-tracking capability and  develop a control policy with better generalization, particularly at higher speeds, we introduce velocity-based rewards. By incorporating a velocity-related term into the reward function, we aim to overcome the limitations of fixed rewards, allowing the robot to adapt its motion patterns to varying commanded velocities.


We conducted a series of comprehensive evaluation experiments in the Humanoid-Gym \cite{gu2024humanoid} using \textit{XBot-L}. The results, which demonstrate accurate velocity tracking at 3.5 m/s, showed the effectiveness of the KSLC method. In additional sim-to-sim validation experiments, the KSLC maintained consistent or even superior performance in a simulator with more realistic physical properties, successfully tracking a wider range of commanded velocities. This indicates its potential for deployment in real-world scenarios. The contributions of this work are summarized as follows: 
\begin{itemize}
    \item This study develops the KSLC, a DRL-based humanoid locomotion control method enhanced with kinodynamic priors, improving locomotion stability in humanoid robots.
    \item Velocity-related reward functions are proposed to avoid the drawback of conservative control policy caused by fixed rewards. Curriculum learning strategy increases stride frequency while maintaining stability, enabling higher locomotion velocities.
    \item Extensive experiments, including sim-to-sim validation, demonstrate that the KSLC can precisely control the humanoid robot's velocity tracking, achieving a top speed of 3.5 m/s.
\end{itemize}

%

\section{Related Work}
\subsection{DRL-based Humanoid Locomotion Control}
In recent years, there has been a significant increase in research on legged robot locomotion utilizing DRL methods. Given the complex structure and high manufacturing costs of humanoid robots, conducting trial-and-error experiments directly in the real world is both expensive and risky. Compared to quadrupedal robots, humanoid robots face additional challenges due to their higher center-of-mass (CoM) and greater number of joints, which complicates DRL-based control.

Radosavovic et al. \cite{radosavovic2024real} proposed a learning-based approach for controlling humanoid robots using a causal transformer model that predicts the robot's next action based on its history of proprioceptive observations and actions. Zhang et al. \cite{zhang2024whole} tackled the complexity of humanoid locomotion by introducing an adversarial motion prior-based imitation learning framework. Cheng et al. \cite{cheng2024expressive} developed an approach that enhances upper-body control by closely following reference motions, while relaxing constraints on the lower body to improve robustness and adaptability. Taylor et al. \cite{9561591} presented a DRL method to teach bipedal robots human-like movements directly from motion capture data, bridging the gap between simulation and reality. Their method incorporates motion re-targeting during training and employs domain randomization techniques to address differences in joint configurations and physical systems.

However, to the best of our knowledge, existing methods have not adequately addressed the impact of upper-body movements on whole-body kinodynamics in achieving accurate and stable locomotion control in humanoid robots. Furthermore, the ability to achieve precise high-velocity tracking in humanoid robot locomotion has not yet been reported. Drawing inspiration from Raibert \cite{raibert1986legged}, we analyze how humans maintain stability while moving at high speeds and propose a DRL method that incorporates kinodynamic priors for humanoid robot locomotion control, utilizing only a limited number of joints.


\subsection{Angular Momentum-based Humanoid Control}
The angular momentum-based control method was introduced decades ago to enhance walking efficiency and performance, specifically by minimizing angular momentum around the CoM during the support phase \cite{raibert1986legged}. Although methods such as DRL and MPC have become mainstream, it is still valuable to consider real-world physical effects. Several studies have analyzed angular momentum and other physical quantities during human motion from a biomechanical perspective \cite{popovic2004angular, martelli2013angular, bennett2010angular}. Following these biomechanical studies, some methods combining MPC with angular momentum analysis have been developed. Kojio et al. \cite{9126153} proposed a gait modification method that considers not only the position of the steps but also their timing and angular momentum. Specifically, this method represented the walkable area as a convex hull and calculated the robot's foot placement, step timing, and angular momentum for each step. Ding et al. \cite{ding2021nonlinear} introduced a Nonlinear Model Predictive Control (NMPC) framework integrating multiple balance strategies, including step location adjustment, CoM height variation, and angular momentum adaptation, for maintaining stability under dynamic disturbances. However, methods that combine the angular momentum-based control method with DRL have not been explored.
\begin{figure}[t]
    \centering
    \includegraphics[width=0.4\textwidth]{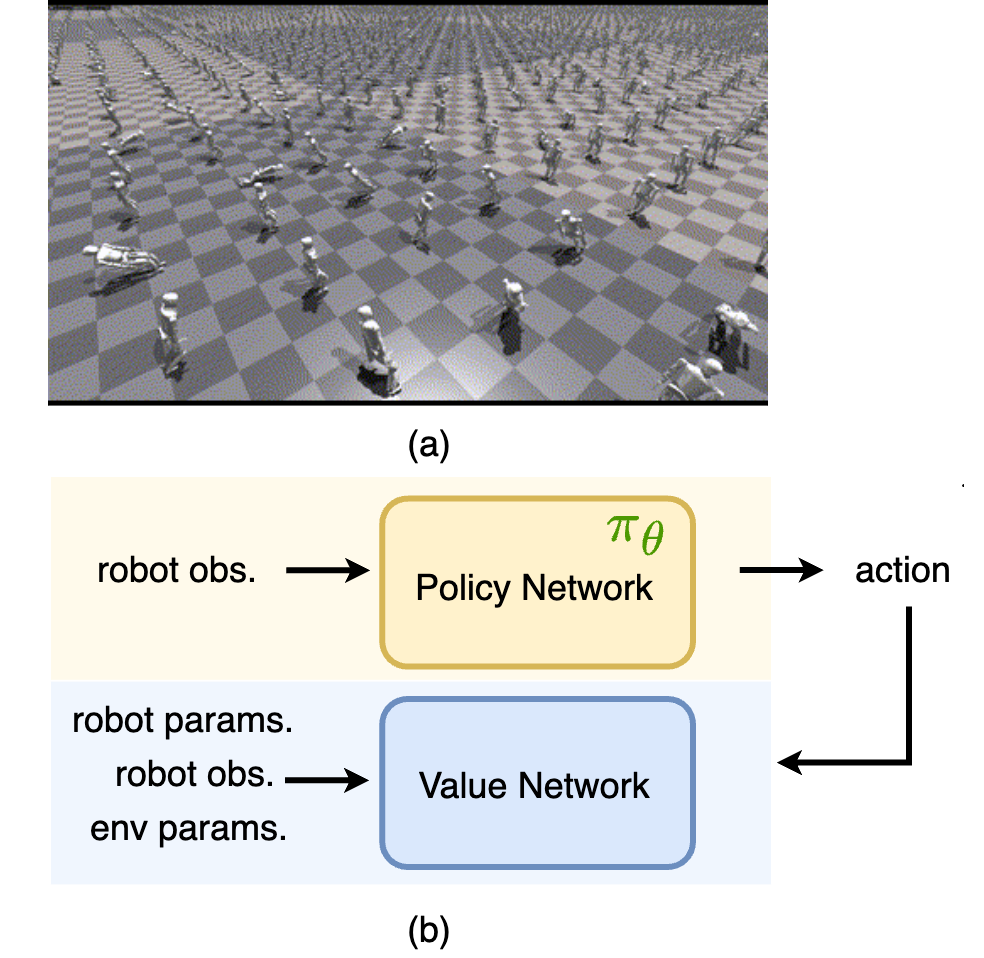}
    \caption{(a) Training environments: Humanoid-Gym \cite{gu2024humanoid}. (b) Actor-Critic network structure.}
    \label{fig:network}
\end{figure}

\section{Preliminaries}

\subsection{Problem Formulation}
The primary objective of this work is to achieve a precise control of locomotion velocity in humanoid robots, particularly in high-speed scenarios. Specifically, we aim to find an optimal control policy $\pi^\ast$ that minimizes the difference between the robot's actual velocity and the commanded velocity, as defined by (\ref{eq1}). Here, $v^{\pi}$ represents the velocity of the humanoid robot under the control of policy $\pi$, and $v^\text{cmd}$ is the commanded velocity.

\begin{equation}
    \begin{aligned}
        \pi^\ast = \underset{\pi}{\text{argmin}} \left( \Vert v^{\pi} - v^\text{cmd} \Vert_1 \right)
    \end{aligned}
    \label{eq1}
\end{equation}

This problem can be formulated as a sequential decision-making problem for achieving the precise velocity control. At time $t$, the input $x_t\in\mathcal{X}$ of the agent contains the current and history information of the humanoid robot, such as periodic clock signals, proprioceptive sensor data, and commanded velocities. The action $a_t \in \mathcal{A}$ refers to the target joint positions of the robot. Given $x_t$, the humanoid robot executes the action $a_t$ based on the current control policy $\pi$. Upon receiving a new observation at time $t+1$, the agent incorporates this information into the history, updates the input to $x_{t+1}$, and receives a reward $r_t$ from the reward function.

We employ DRL to train a deep neural network (DNN), $\pi_\theta$, to approximate the optimal control policy $\pi^\ast$, where $\theta$ represents the parameters of the model. These parameters are optimized by maximizing the expected total rewards during training. The DRL algorithm used in this study is Proximal Policy Optimization (PPO) \cite{schulman2017proximal}, chosen for its efficiency and suitability for parallel computation.

\begin{figure}[t]
    \centering
    \includegraphics[width=0.4\textwidth]{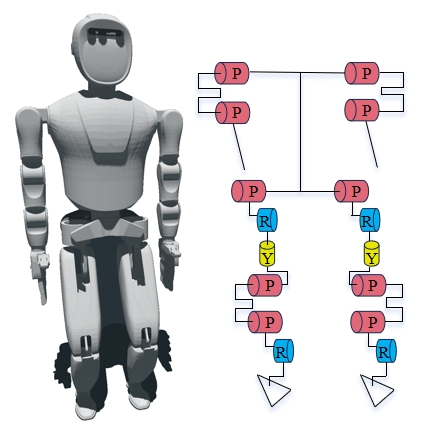}
    \caption{Humanoid robot \textit{XBot-L} and Degrees-of-Freedom (DoF) configuration \cite{gu2024humanoid}.}
    \label{fig:joint}
\end{figure}

\subsection{Kinodynamics Priors}
Humanoid robots have numerous movable joints, which, if left unrestricted, can often result in instability during motion. In kinodynamics, maintaining stability requires minimizing the overall angular momentum of the robot to prevent such instability. Rather than fixing the upper-body joints, we encourage their use to counterbalance the angular momentum generated by the legs. Specifically, we allow the shoulder pitch and elbow pitch joints of \textit{XBot-L}, as shown in Fig. \ref{fig:joint}, to generate angular momentum that counteracts the leg swing.

\section{Methods}
\begin{figure}[t]
    \centering
    \includegraphics[width=0.48\textwidth]{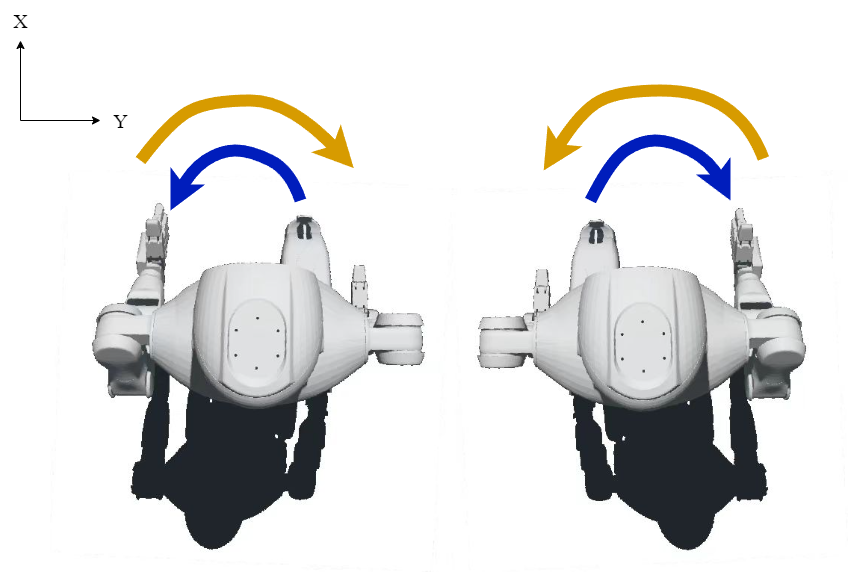}
    \caption{Angular momentum of the arms and legs under two typical locomotion status. Yellow arrows and blue arrows indicate the direction of angular momentum generated in the arms and legs, respectively.}
    \label{fig:angular}
\end{figure}

Recent studies have demonstrated that humanoid robots can maintain stable walking at low velocities \cite{radosavovic2024real, gu2024humanoid, zhang2024whole}. However, the lack of consideration for the effects of angular momentum on whole-body stability during walking and other dynamic tasks often leads to unnatural movements or even falls when tracking higher velocities \cite{sferrazza2024humanoidbench}. To address these instability challenges in high-velocity locomotion, we combine DRL with kinodynamics and develop an angular momentum-based approach. This approach effectively utilizes the upper-body joints to counterbalance the angular momentum generated by the lower body at higher commanded velocities \cite{hinrichs1990whole, hinrichs1987upper}, mitigating the negative impact seen when only lower-body joints are engaged. As a result, the overall stability of the humanoid robot is significantly enhanced, as illustrated in Fig. \ref{fig:angular}.

At varying commanded velocities, especially at high speeds, fixed reward functions gradually become inadequate for the demands of rapid locomotion in humanoid robots, as they typically focus on a single motion pattern, such as walking. We observed that fixed reward functions lead the robot to struggle with tracking higher velocities, often defaulting to slower walking behaviors. This failure is due to insufficient rewards or early terminations, prompting the robot to adopt overly conservative strategies at higher speeds. To address this issue, we propose velocity-related reward functions designed to improve velocity-tracking performance across a wide range of commanded speeds.

The total reward is a weighted sum of three categories of rewards: 1) angular momentum-based rewards, 2) velocity-related rewards, and 3) general rewards. Specifically, the expected total reward at time $t$ is given by:
\begin{equation}
    \begin{aligned}
        r_t &\ = \alpha^a r^a_t + \alpha^v r^v_t + \alpha^c r^c_t
    \end{aligned}
    \label{eq2}
\end{equation}
where $\alpha^a$, $\alpha^v$, and $\alpha^c$ are coefficients that control the weighting of each reward component in the total reward function. In our work, $\alpha^a$ and $\alpha^v$ are set to 0.05 and 0.1, respectively, with the remaining coefficients following those used in \cite{gu2024humanoid}.

\subsection{Angular Momentum-based Reward}
According to the kinodynamics of humanoid robot locomotion, the angular momentum generated by the swinging motion of the robot's legs is significant and cannot be ignored. Moreover, this angular momentum increases with the robot's velocity, especially in the yaw direction \cite{otani2018upper}. The total angular momentum is the sum of the angular momentum produced by the robot's individual links during locomotion, as described in (\ref{eq3}),
\begin{equation}
    \begin{aligned}
        \mathbf{L}_{\text{total}} = \sum_{i=1}^{n} \left( \mathbf{c}_i \times (m_i \dot{\mathbf{c}}_i) + \mathbf{I}_i \mathbf{\omega}_i \right)
        \label{eq3}
    \end{aligned}
\end{equation}
where $\mathbf{L}_{\text{total}}$ represents the total angular momentum, $n$ is the number of links, $\mathbf{I}_i$ is the inertia tensor of the $i^{th}$ link's CoM, $\mathbf{\omega}_i$ denotes the angular velocity, and $\mathbf{c}_i$ and $\dot{\mathbf{c}}_i$ represent the position and velocity of the $i^{th}$ link's CoM relative to the whole-body CoM, respectively..

The substantial angular momentum generated during high-velocity locomotion presents a considerable challenge to precise control, especially in dynamic movements, where high stability is essential. To maintain postural equilibrium, we designed a reward function to minimize angular momentum for more stable and efficient locomotion. The angular momentum-based reward is defined in (\ref{eq4}),
\begin{equation}
    \begin{aligned}
        r^a_t = \text{clip}(-\exp\left(\Vert A^a_t \Vert_2 \right), c_1, c_2)
        \label{eq4}
    \end{aligned}
\end{equation}
where $\Vert A^a_t \Vert_2$ represents the $L_2$ norm of the angular momentum at time $t$. The function $\text{clip}(\psi, c_1, c_2)$ constrains the output of the function $\psi$ to mitigate extreme values. Here, $\psi$ signifies any function that necessitates the clipping operation, with $c_1$ and $c_2$ indicating the lower and upper bounds of the output value, respectively.

\subsection{Velocity-related Reward}
In locomotion tasks, reward functions play a critical role in shaping the behavior of humanoid robots. Previous studies often rely on fixed reward functions, which limits the humanoid robots' ability to accurately track velocities across a broad range of commanded velocities. To enhance the generalization capabilities of humanoid robots, we propose velocity-related reward functions that promote the learning of more dynamically adaptive behaviors. These functions enable precise control even over a wide range of commanded velocities. The rewards are categorized as follows: 1) base height reward, 2) feet clearance reward, and 3) joint position reward. In our approach, the coefficients $\alpha^b$, $\alpha^f$, and $\alpha^{p}$ are set to 0.2, 1.0, and 1.6, respectively.
\begin{equation}
    \begin{aligned}
        r^v_t = \alpha^b r^b_t + \alpha^f r^f_t + \alpha^{p} r^p_t
    \end{aligned}
    \label{eq5}
\end{equation}

\subsubsection{Base Height}
In human locomotion, the CoM height is automatically adjusted in preparation for subsequent steps \cite{ernst2019humans}. In contrast, humanoid robots typically maintain a fixed base height, which restricts the range of motion patterns they can learn during training. To overcome this limitation, we develop a reward that enables dynamic adjustment of the robot’s base height during training. This approach significantly enhances the robot’s responsiveness to changes in velocity and improves stability at higher velocities. The base height reward is calculated as follows:
\begin{equation}
    \begin{aligned}
        r^b_t = \exp\left(-\beta^b_t \Vert h^b_t - \gamma^b_t \hat{h^b_t} \text{g}(v^\text{cmd}_t, v^{\text{max}}_t) \Vert_2 \right)
    \label{eq6}
    \end{aligned}
\end{equation} 
where $h^b_t$ and $\hat{h^b_t}$ denote the current and target base heights at time $t$, respectively. The coefficients $\beta^b_t$ and $\gamma^b_t$ are set to 100 and 0.05 in our work. The function $\text{g}(v^\text{cmd}_t, v^\text{max}_t)$ is defined as $v^\text{cmd}_t$ divided by $v^\text{max}_t$, where $v^\text{cmd}_t$ is the commanded velocity and $v^\text{max}_t$ is the maximum velocity of curriculum at time $t$.

\subsubsection{Feet Clearance}
Intuitively, maintaining appropriate feet clearance and ensuring an optimal distance between the swinging leg and the ground can help prevent tripping, thus enhancing the humanoid robot's ability to navigate challenging terrain, particularly in environments with obstacles. To facilitate this, we propose a reward function that incorporates both feet clearance and locomotion velocity. The reward function is defined as follows:
\begin{equation}
    \begin{aligned}
        r^f_t = \xi(t) \cdot \left(h^f_t - \hat{h^f_t} \text{g}(v^\text{cmd}_t, v^\text{max}_t) \right)
    \label{eq7}
    \end{aligned}
\end{equation}
where $h_t^f$ and $\hat{h_t^f}$ denote the current and target heights of feet, respectively. Following \cite{gu2024humanoid}, $\xi(t)$ represents a swing mask function that characterizes the swing phase of the gait cycle.

\subsubsection{Joint Position}
In high-velocity locomotion control tasks, accurate velocity tracking is influenced by stride length. By examining the relationship between joint angles and locomotion velocity of the humanoid robot, we develop a reward to improve velocity tracking by facilitating an appropriate increase in joint angles, thereby enabling longer strides. The reward is expressed as follows:
\begin{equation}
    \begin{aligned}
        r^p_t = & \ \exp(\beta^p_t \Vert (\hat{\theta^p_t} - \theta^p_t) \exp(-\dot{\theta}^p_t) \Vert_2) \\
        & \ - \gamma^p_t \text{clip}(\Vert (\hat{\theta^p_t} - \theta^p_t) \exp(-\dot{\theta^p_t})\Vert_2 , c_1, c_2)
    \end{aligned}
    \label{eq8}
\end{equation}
where $\theta^p_t$ and $\hat{\theta^p_t}$ denote the current and target joint positions, respectively. The term $\exp(-\dot{\theta}^p_t)$ is a velocity-related coefficient that adjusts the importance of the velocity-related term. The parameters $\beta^p_t$ and $\gamma^p_t$ are the tolerances for joint position tracking error, which are set to $-2$ and $0.2$ in this work. The constants $c_1$ and $c_2$, used to truncate excessively large values, are set to $0$ and $0.5$, respectively.

\subsection{Curriculum Learning Strategy}
\subsubsection{Velocity Curriculum}
During DRL training, the humanoid robot learns a policy by tracking various commanded velocities to develop velocity tracking capabilities. However, as the range of commanded velocities expands, the humanoid robot faces challenges in accurately tracking high-velocity commands during rapid locomotion. This difficulty arises because the humanoid robot may struggle to earn rewards if the early training is dominated by high-velocity commands, potentially leading to failure in rapid locomotion tasks \cite{margolis2024rapid}. To address this issue, we trained the humanoid robot using low-velocity commands initially and progressively expanded the velocity range as per the curriculum \cite{bengio2009curriculum}.

In addition, we employ a reward-based update rule rather than a fixed one to automatically adjust the curriculum during training, eliminating the need for manual adjustments \cite{lee2020learning, matiisen2019teacher}. The update rule $f^v_t$ is defined as follows:
\begin{equation}
    [v^\text{min}_{t+1}, v^\text{max}_{t+1}] \gets 
    \begin{cases} 
        f^v_t([v^\text{min}_t, v^\text{max}_t], r^\text{trk}_t), & r^\text{trk}_t \geq \lambda r^\text{trk, max}_t                  \\
        [v^\text{min}_t, v^\text{max}_t], & \text{otherwise.}
    \end{cases}
    \label{eq9}
\end{equation}
where $[v^\text{min}_t, v^\text{max}_t]$ is the minimum and maximum values of the current linear velocity command in the forward direction, $r^\text{trk}_t$ and $r^\text{trk, max}_t$ are the linear velocity tracking reward and the maximum achievable linear velocity tracking reward at time $t$. If the threshold $\lambda r^\text{trk, max}_t$ is satisfied, the velocity command interval will be updated to expand the sampling range of the velocity command. In this work, $\lambda$ is set to 0.8.

\begin{figure}[t]
    \centering
    \includegraphics[width=0.48\textwidth]{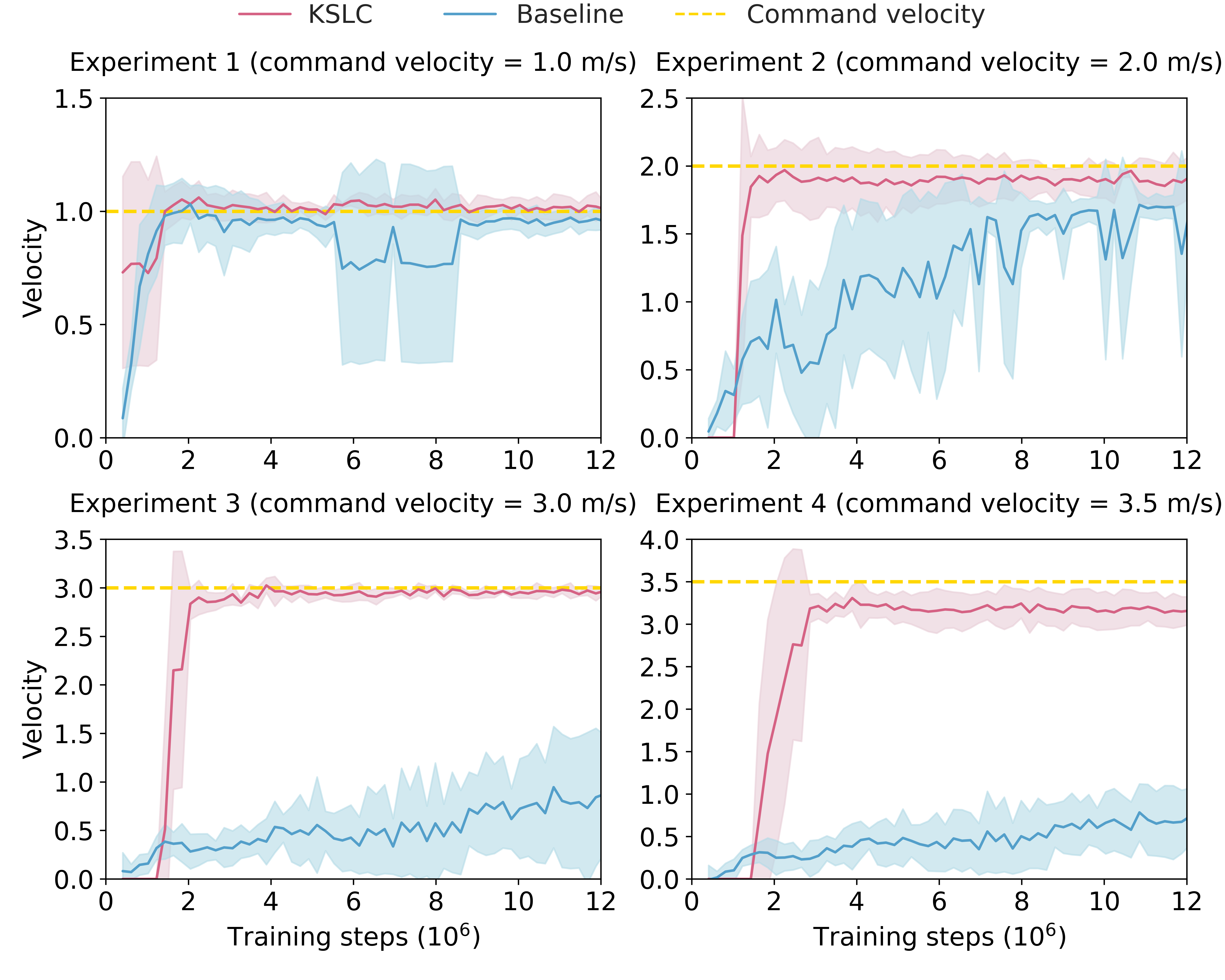}
    \caption{Learning curves of the humanoid robot \textit{XBot-L} trained with the KSLC and the baseline method at different commanded velocities.}
    \label{fig:Checkpoint}
\end{figure}

\subsubsection{Cycle Time Curriculum}
The gait cycle determines the stride frequency of the humanoid robot and is crucial for achieving high-speed locomotion. The fixed gait cycle limits the robot's ability to track higher commanded velocities, potentially leading to failure in maintaining the commanded velocities or causing the humanoid robot to fall. To address this limitation, we propose a curriculum learning strategy focused on cycle time, which encourages the humanoid robot to progressively shorten leg movement cycle times as commanded velocity increases, thereby enhancing stride frequency. The update rule $f^c_t$ is shown as follows:
\begin{equation}
    \tau^c_{t+1} \gets 
    \begin{cases} 
        f^c_t(\tau^c_t, r^\text{trk}_t), & r^\text{trk}_t \geq \lambda r^\text{trk, max}_t \\
        \tau^c_t, & \text{otherwise.}
    \end{cases}
    \label{eq10}
\end{equation}
where $\tau^c_t$ is the cycle time at time $t$, the update condition follows the same as (\ref{eq9}). Our function encourages the robot to adjust and modulate the gait cycle duration in response to varying commanded velocities, ensuring the optimal stride frequency is achieved.

\section{Experiments}
In this section, we illustrate the training procedure conducted in the Humanoid-Gym simulator and the subsequent sim-to-sim validation in a high-fidelity simulator. We trained a policy using the KSLC method within the Humanoid-Gym simulation environment \cite{gu2024humanoid} on a workstation equipped with an NVIDIA 4090 GPU. To further validate the effectiveness of KSLC, we performed zero-shot sim-to-sim experiments in a high-fidelity simulator.

\begin{figure}[t]
    \centering
    \includegraphics[width=0.48\textwidth]{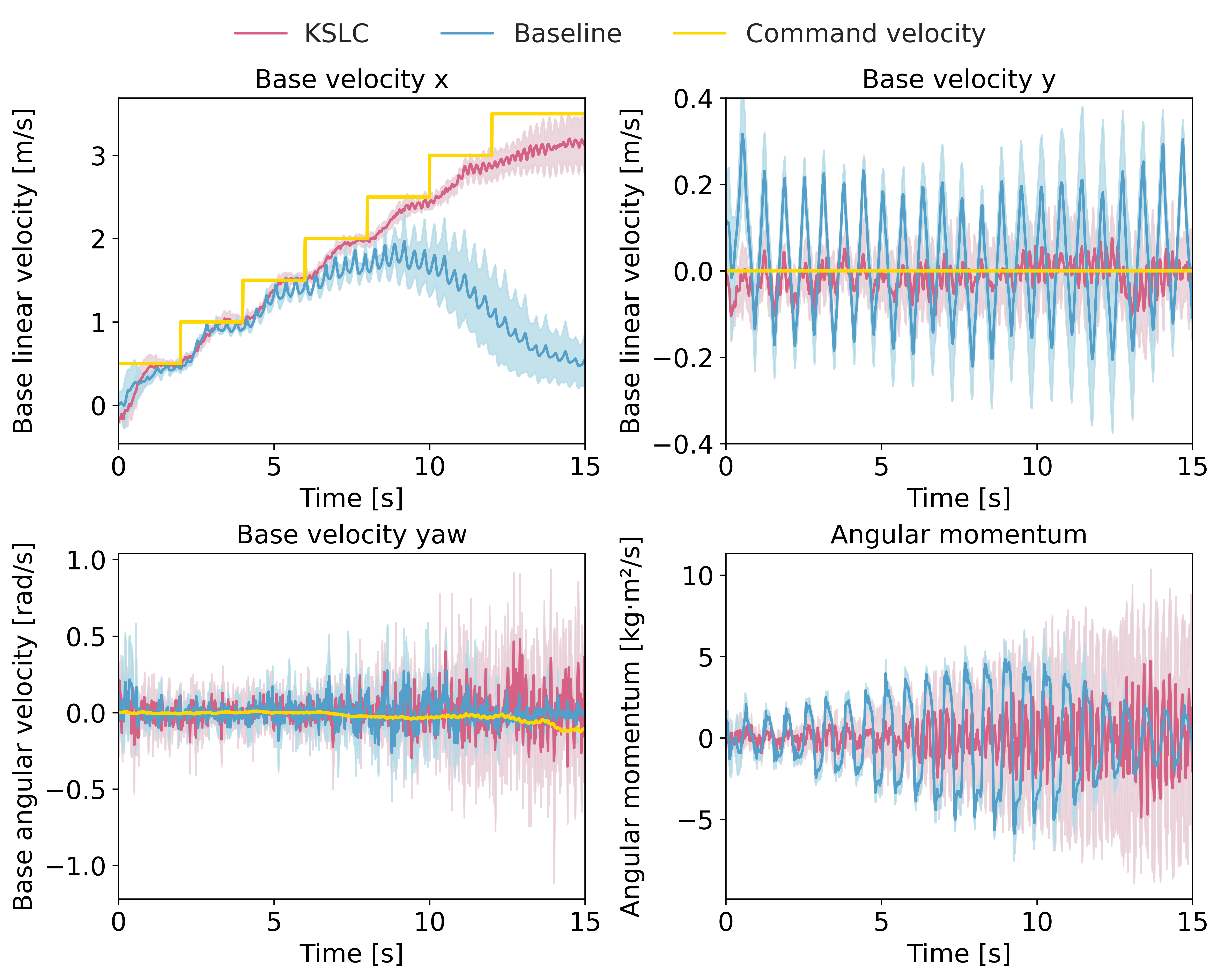}
    \caption{Performance of the humanoid robot \textit{XBot-L} across various metrics when tracking different commanded velocities, using the policy trained with the KSLC and baseline method in the Humanoid-Gym \cite{gu2024humanoid}.}
    \label{fig:play}
\end{figure}

\subsection{Training Details}
\subsubsection{Training Configuration}
We released the left and right shoulder pitch joints, as well as the left and right elbow pitch joints of \textit{XBot-L}. The action space is defined as $a_t \in \mathbb{R}^{16}$, representing the target joint positions for the Proportional-Derivative (PD) controller. In addition, we applied curriculum learning strategies to both the commanded velocity and the gait cycle time during training. The maximum commanded velocity was set to 3.5 m/s, with the sampling range adjusted in increments of 0.5 m/s after each update. For the gait cycle time, each update reduced the cycle time to 95\% of its previous value, with a minimum period set at 0.48 s. Detailed information on the observation and privileged observation spaces are provided in Table \ref{tab:observation}.

We trained our policy using five different random seeds. The detailed hyperparameter settings for the algorithms and environment are listed in Table \ref{tab:hyperparameters}. The KSLC method significantly improves the policies' ability to accurately track commanded velocities, as illustrated in Fig. \ref{fig:Checkpoint}. We quantified the tracking performance by averaging the humanoid robot's velocity during tests at each commanded velocity. If the robot fell during testing, the average velocity was recorded as 0 m/s. The results demonstrated that the robot controlled by the KSLC method accurately followed the commanded velocities while maintaining greater stability, as evidenced by smaller fluctuations in the zero-commanded velocity directions. In contrast, the baseline method failed to track high commanded velocities, resorting instead to conservative low-velocity locomotion, as shown in Fig. \ref{fig:play}.

\subsubsection{Domain Randomization}
To minimize the gap between simulators and the real world, addressing issues such as modeling discrepancies and unexpected disturbances, we employed domain randomization to train a more robust control policy \cite{tobin2017domain}. The effectiveness of the policy trained using domain randomization was validated in a high-fidelity simulator, as detailed in the following sections.

\begin{table}[t]
\renewcommand{\arraystretch}{1.3}
\caption{Summary of Hyperparameters}
\centering
\label{tab:hyperparameters}
\begin{tabular}{cc}
\hline
\hline
\textbf{Parameter} & \textbf{Value} \\
\hline
Number of Environments & 4096 \\
Mini-batch size & 61440 \\
Discount Factor & 0.994 \\
GAE discount factor & 0.9 \\
Entropy Regularization Coefficient & 0.001 \\
Angular momentum factor & -0.05 \\
Learning rate & 1e-5 \\
Frame Stack of Single Observation & 15 \\
Frame Stack of Single Privileged Observation & 3 \\
Number of Single Observation & 59 \\
Number of Single Privileged Observation & 89 \\
\hline
\hline
\end{tabular}
\end{table}

\begin{table}[t]
\caption{Summary of Observation Space}
\centering
\label{tab:observation}
\begin{tabular}{cccc}
\hline
\hline
\textbf{Components} & \textbf{Dims} & \textbf{Observation} & \textbf{State} \\
\hline
Clock Input \((\sin(t), \cos(t))\) & 2 & \checkmark & \checkmark \\
Commands \((\dot{P}_{x, y, \gamma})\) & 3 & \checkmark & \checkmark \\
Joint Position $(\theta)$ & 16 & \checkmark & \checkmark \\
Joint Velocity $(\dot{\theta})$ & 16 & \checkmark & \checkmark \\
Angular Velocity $(\dot{P}^b_{\alpha \beta \gamma})$  & 3 & \checkmark & \checkmark \\
Euler Angle $(P^b_{\alpha \beta \gamma})$  & 3 & \checkmark & \checkmark \\
Last Actions $(a_{t - 1})$ & 16 & \checkmark & \checkmark \\
Frictions & 1 & & \checkmark \\
Body Mass & 1 & & \checkmark \\
Base Linear Velocity & 3 & & \checkmark \\
Push Force & 2 & & \checkmark \\
Push Torques & 3 & & \checkmark \\
Tracking Difference & 16 & & \checkmark \\
Periodic Stance Mask & 2 & & \checkmark \\
Feet Contact detection& 2 & & \checkmark \\
\hline
\hline
\end{tabular}
\end{table}

\subsection{Sim-to-sim Validation}
To further evaluate the performance of the KSLC, we conducted sim-to-sim experiments, as access to a physical humanoid robot for sim-to-real testing was unavailable. The disparity between low-fidelity and high-fidelity simulators can be considered analogous to the gap between simulation and real-world conditions. Transferring a policy learned in a low-fidelity simulator to a high-fidelity simulator is known as sim-to-sim transfer \cite{gu2024humanoid}. As demonstrated in \cite{gu2024humanoid}, agents that perform well in sim-to-sim transfer often keep satisfactory performance in sim-to-real transfer. Following this approach, MuJoCo was employed as the high-fidelity simulator in our sim-to-sim experiments. A video of the experiment is available at \url{https://youtu.be/uVT8Up8vAKc} and in the supplementary materials.

\begin{figure}[t]
    \centering
    \includegraphics[width=0.40\textwidth]{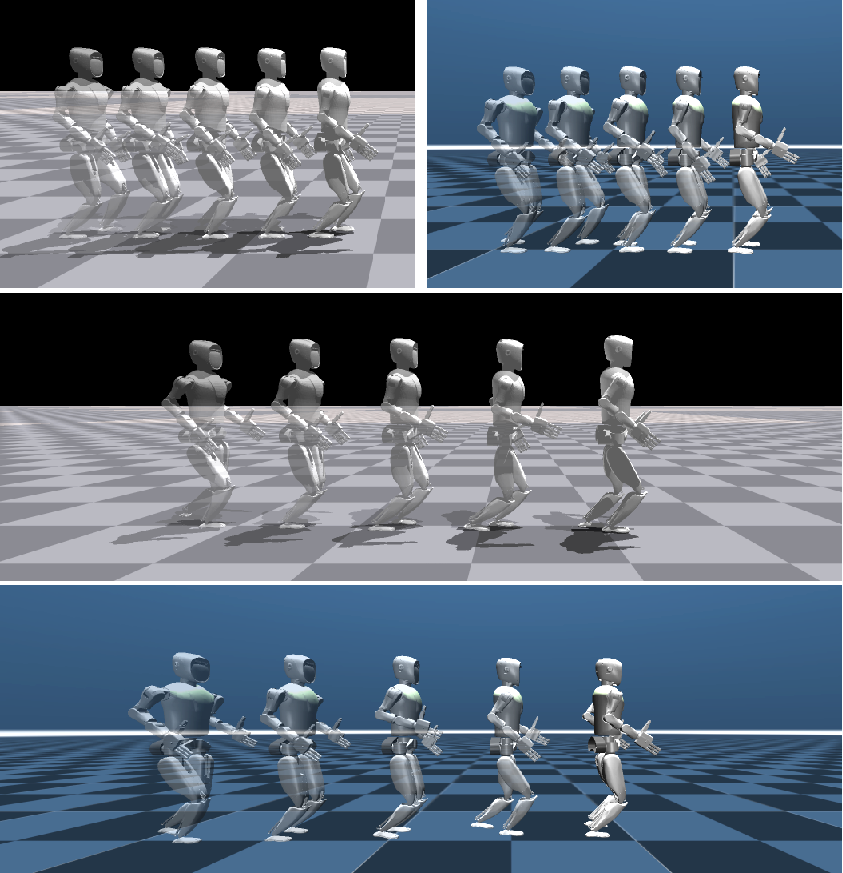}
    \caption{Different gaits at 2.0 m/s, 3.5 m/s commanded velocities in the Humanoid-Gym and high-fidelity simulator. The interval between captured photos is 0.25 s, and the length of each grid is 1.0 m. Our method shows the relationship between gait and velocity in both simulation environments. The video of the experiment is available at \url{https://youtu.be/uVT8Up8vAKc} and in the supplementary material.}
    \label{fig:vel-gait}
\end{figure}

The sim-to-sim transfer results demonstrate that the policy trained with the KSLC in the Humanoid-Gym maintained consistent or even superior performance in the MuJoCo. The \textit{XBot-L} successfully tracked a commanded velocity of 4.0 m/s without falling. As illustrated in Fig. \ref{fig:vel-gait}, the arm movement patterns remained consistent with those observed in the Humanoid-Gym. Furthermore, the humanoid robot utilizing the KSLC method exhibited distinct behaviors at different commanded velocities, supporting its anticipated stability during high-speed locomotion.

\section{Conclusion}
In this work, we have proposed the KSLC, a method for achieving stable locomotion for humanoid robots by integrating DRL with kinodynamic priors, based on our analysis of the factors that enable humans to move both stably and quickly. By incorporating these priors, the KSLC method has facilitated the exploration of effective locomotion behaviors, with improved stability and enhanced velocity tracking during rapid locomotion. Extensive experimental results have demonstrated that the KSLC method enables humanoid robots to track a broader range of commanded velocities, up to 3.5 m/s, with reduced fluctuations compared to the baseline. In sim-to-sim validation, the humanoid robot enabled by the KSLC has consistently maintained its performance, exhibiting locomotion behaviors similar to those observed in the training environment. Inspired by the effectiveness of the KSLC, we plan to incorporate more human-inspired priors to further enhance the performance of humanoid robots.

\addtolength{\textheight}{-6cm}


%



\ifCLASSOPTIONcaptionsoff
  \newpage
\fi



%




\bibliographystyle{IEEEtran}
\bibliography{ref}

\begin{thebibliography}{10}
\providecommand{\url}[1]{#1}
\csname url@rmstyle\endcsname
\providecommand{\newblock}{\relax}
\providecommand{\bibinfo}[2]{#2}
\providecommand\BIBentrySTDinterwordspacing{\spaceskip=0pt\relax}
\providecommand\BIBentryALTinterwordstretchfactor{4}
\providecommand\BIBentryALTinterwordspacing{\spaceskip=\fontdimen2\font plus
\BIBentryALTinterwordstretchfactor\fontdimen3\font minus \fontdimen4\font\relax}
\providecommand\BIBforeignlanguage[2]{{%
\expandafter\ifx\csname l@#1\endcsname\relax
\typeout{** WARNING: IEEEtran.bst: No hyphenation pattern has been}%
\typeout{** loaded for the language `#1'. Using the pattern for}%
\typeout{** the default language instead.}%
\else
\language=\csname l@#1\endcsname
\fi
#2}}

\bibitem{kang2023rl+}
D.~Kang, J.~Cheng, M.~Zamora, F.~Zargarbashi, and S.~Coros, ``Rl+ model-based control: Using on-demand optimal control to learn versatile legged locomotion,'' \emph{IEEE Robotics and Automation Letters}, 2023.

\bibitem{rudin2022learning}
N.~Rudin, D.~Hoeller, P.~Reist, and M.~Hutter, ``Learning to walk in minutes using massively parallel deep reinforcement learning,'' in \emph{Conference on Robot Learning}.\hskip 1em plus 0.5em minus 0.4em\relax PMLR, 2022, pp. 91--100.

\bibitem{hoeller2024anymal}
D.~Hoeller, N.~Rudin, D.~Sako, and M.~Hutter, ``Anymal parkour: Learning agile navigation for quadrupedal robots,'' \emph{Science Robotics}, vol.~9, no.~88, p. eadi7566, 2024.

\bibitem{margolis2024rapid}
G.~B. Margolis, G.~Yang, K.~Paigwar, T.~Chen, and P.~Agrawal, ``Rapid locomotion via reinforcement learning,'' \emph{The International Journal of Robotics Research}, vol.~43, no.~4, pp. 572--587, 2024.

\bibitem{cheng2024expressive}
X.~Cheng, Y.~Ji, J.~Chen, R.~Yang, G.~Yang, and X.~Wang, ``Expressive whole-body control for humanoid robots,'' in \emph{Robotics: Science and Systems}, 2024.

\bibitem{radosavovic2024real}
I.~Radosavovic, T.~Xiao, B.~Zhang, T.~Darrell, J.~Malik, and K.~Sreenath, ``Real-world humanoid locomotion with reinforcement learning,'' \emph{Science Robotics}, vol.~9, no.~89, p. eadi9579, 2024.

\bibitem{gu2024humanoid}
X.~Gu, Y.-J. Wang, and J.~Chen, ``Humanoid-gym: Reinforcement learning for humanoid robot with zero-shot sim2real transfer,'' \emph{arXiv preprint arXiv:2404.05695}, 2024.

\bibitem{sferrazza2024humanoidbench}
C.~Sferrazza, D.-M. Huang, X.~Lin, Y.~Lee, and P.~Abbeel, ``Humanoidbench: Simulated humanoid benchmark for whole-body locomotion and manipulation,'' in \emph{Robotics: Science and Systems}, 2024.

\bibitem{otani2018upper}
T.~Otani, K.~Hashimoto, S.~Miyamae, H.~Ueta, A.~Natsuhara, M.~Sakaguchi, Y.~Kawakami, H.-O. Lim, and A.~Takanishi, ``Upper-body control and mechanism of humanoids to compensate for angular momentum in the yaw direction based on human running,'' \emph{Applied Sciences}, vol.~8, no.~1, p.~44, 2018.

\bibitem{raibert1986legged}
M.~H. Raibert, \emph{Legged robots that balance}.\hskip 1em plus 0.5em minus 0.4em\relax MIT press, 1986.

\bibitem{zhang2024whole}
Q.~Zhang, P.~Cui, D.~Yan, J.~Sun, Y.~Duan, A.~Zhang, and R.~Xu, ``Whole-body humanoid robot locomotion with human reference,'' \emph{arXiv preprint arXiv:2402.18294}, 2024.

\bibitem{9561591}
M.~Taylor, S.~Bashkirov, J.~F. Rico, I.~Toriyama, N.~Miyada, H.~Yanagisawa, and K.~Ishizuka, ``Learning bipedal robot locomotion from human movement,'' in \emph{2021 IEEE International Conference on Robotics and Automation}, 2021, pp. 2797--2803.

\bibitem{popovic2004angular}
M.~Popovic, A.~Hofmann, and H.~Herr, ``Angular momentum regulation during human walking: biomechanics and control,'' in \emph{2004 IEEE International Conference on Robotics and Automation}, 2004, pp. 2405--2411.

\bibitem{martelli2013angular}
D.~Martelli, V.~Monaco, L.~B. Luciani, and S.~Micera, ``Angular momentum during unexpected multidirectional perturbations delivered while walking,'' \emph{IEEE Transactions on Biomedical Engineering}, vol.~60, no.~7, pp. 1785--1795, 2013.

\bibitem{bennett2010angular}
B.~C. Bennett, S.~D. Russell, P.~Sheth, and M.~F. Abel, ``Angular momentum of walking at different speeds,'' \emph{Human movement science}, vol.~29, no.~1, pp. 114--124, 2010.

\bibitem{9126153}
Y.~Kojio, Y.~Omori, K.~Kojima, F.~Sugai, Y.~Kakiuchi, K.~Okada, and M.~Inaba, ``Footstep modification including step time and angular momentum under disturbances on sparse footholds,'' \emph{IEEE Robotics and Automation Letters}, vol.~5, no.~3, pp. 4907--4914, 2020.

\bibitem{ding2021nonlinear}
J.~Ding, C.~Zhou, S.~Xin, X.~Xiao, and N.~G. Tsagarakis, ``Nonlinear model predictive control for robust bipedal locomotion: exploring angular momentum and com height changes,'' \emph{Advanced Robotics}, vol.~35, no.~18, pp. 1079--1097, 2021.

\bibitem{schulman2017proximal}
J.~Schulman, F.~Wolski, P.~Dhariwal, A.~Radford, and O.~Klimov, ``Proximal policy optimization algorithms,'' \emph{arXiv preprint arXiv:1707.06347}, 2017.

\bibitem{hinrichs1990whole}
R.~N. Hinrichs, ``Whole body movement: coordination of arms and legs in walking and running,'' \emph{Multiple muscle systems: biomechanics and movement organization}, pp. 694--705, 1990.

\bibitem{hinrichs1987upper}
R.~N. Hinrichs, ``Upper extremity function in running. ii: Angular momentum considerations,'' \emph{Journal of Applied Biomechanics}, vol.~3, no.~3, pp. 242--263, 1987.

\bibitem{ernst2019humans}
M.~Ernst, M.~G{\"o}tze, R.~Blickhan, and R.~M{\"u}ller, ``Humans adjust the height of their center of mass within one step when running across camouflaged changes in ground level,'' \emph{Journal of biomechanics}, vol.~84, pp. 278--283, 2019.

\bibitem{bengio2009curriculum}
Y.~Bengio, J.~Louradour, R.~Collobert, and J.~Weston, ``Curriculum learning,'' in \emph{Proceedings of the 26th annual international conference on machine learning}, 2009, pp. 41--48.

\bibitem{lee2020learning}
J.~Lee, J.~Hwangbo, L.~Wellhausen, V.~Koltun, and M.~Hutter, ``Learning quadrupedal locomotion over challenging terrain,'' \emph{Science robotics}, vol.~5, no.~47, p. eabc5986, 2020.

\bibitem{matiisen2019teacher}
T.~Matiisen, A.~Oliver, T.~Cohen, and J.~Schulman, ``Teacher--student curriculum learning,'' \emph{IEEE transactions on neural networks and learning systems}, vol.~31, no.~9, pp. 3732--3740, 2019.

\bibitem{tobin2017domain}
J.~Tobin, R.~Fong, A.~Ray, J.~Schneider, W.~Zaremba, and P.~Abbeel, ``Domain randomization for transferring deep neural networks from simulation to the real world,'' in \emph{2017 IEEE/RSJ international conference on intelligent robots and systems}, 2017, pp. 23--30.

\end{thebibliography}

%








\end{document}